# Non-Intrusive Load Monitoring Based on Image Load Signatures and Continual Learning

Olimjon TOIROV

Electrical Engineering Faculty, Tashkent State Technical University, Tashkent, 100095, Uzbekistan.

E-mail: olimjontoirov@gmail.com

Wei Yu *

School of Artificial Intelligence, Wuhan University, Wuhan, 430072, China

Email: yuwei@whu.edu.cn

**Abstract**

Non-Intrusive Load Monitoring (NILM) identifies the operating status and energy consumption of each electrical device in the circuit by analyzing the electrical signals at the bus, which is of great significance for smart power management. However, the complex and changeable load combinations and application environments lead to the challenges of poor feature robustness and insufficient model generalization of traditional NILM methods. To this end, this paper proposes a new non-intrusive load monitoring method that integrates "image load signature" and continual learning. This method converts multi-dimensional power signals such as current, voltage, and power factor into visual image load feature signatures, and combines deep convolutional neural networks to realize the identification and classification of multiple devices; at the same time, self-supervised pre-training is introduced to improve feature generalization, and continual online learning strategies are used to overcome model forgetting to adapt to the emergence of new loads. This paper conducts a large number of experiments on high-sampling rate load datasets, and compares a variety of existing methods and model variants. The results show that the proposed method has achieved significant improvements in recognition accuracy.

**CCS CONCEPTS**

General and reference ~ Document types ~ General conference proceedings

## 1 INTRODUCTION

Non-Intrusive Load Monitoring (NILM) technology identifies the operating status and energy consumption of each electrical device in the circuit by collecting electrical signals only at the main power supply. It has broad application prospects in the fields of smart grid and home energy management [1]. Compared with invasive approach that requires installing independent sensors on device, NILM offers notable benefits in terms of deployment convenience. However, due to diverse power characteristics of electrical appliances and the signal superposition when multiple devices are running, current NILM methods still face limitations in the robustness of feature extraction. Early NILM research mainly relied on manually extracted features and used simple classifiers for identification, but the performance was poor when the state combination was complex.

Recent years, with the development of AI, many studies have begun to apply deep learning to NILM to automatically extract discriminative features. Such methods often require massive labeled data and have limited generalization when migrating to different environments. In response to the lack of labeled data and domain differences, some scholars introduced a self-supervised learning mechanism to pre-train feature extraction networks using unlabeled data to improve the versatility of the model. For example, Chen et al. [2] proposed to first perform self-supervised pre-training on the total power sequence of the target user, and then fine-tune the supervised load classification, thereby significantly improving the cross-domain recognition performance without the target user's sub-device label. Meanwhile, the high-frequency details and multi-dimensional features of the current and voltage signals make it possible to distinguish different electrical appliances. Studies have shown [3] that image features generated using voltage-current (V-I) trajectory curves can be used for load identification and have achieved a high accuracy rate.

However, most of deep learning-based load monitoring methods only identify fixed load types, and lack the ability to adapt to new equipment types or changes in operating conditions after deployment. When the model encounters a new appliance that does not appear in training set, it often cannot correctly identify it, which limits the scalability of NILM in actual scenarios [4]. To solve this problem, continual learning technology has been introduced into the NILM field [5]. Lu et al. [6] proposed a retrainable twin network for load identification. When an unknown appliance is detected, the model is incrementally updated through user annotations to improve the recognition accuracy of the new load. Paknezhad et al. [7] introduced a parallel continual learning strategy (PaRT), which assigns task-specific subnetworks and simultaneously trains on new and previous tasks using shared and independent modules, thereby improving representation generalizability while preventing catastrophic forgetting.

In summary, how to extract discriminative feature from complex signals, and how to improve generalization after deployment are key issues that need to be urgently addressed in the research of non-intrusive load monitoring. Therefore, we propose a non-intrusive load monitoring method based on image load signature and continual learning. The core idea is to construct the "image signature" feature of electrical appliance operation by fusing multimodal signals such as current, voltage and power factor, and use convolutional neural network to achieve effective classification of parallel multi-devices; and combine self-supervised pre-training and continual online learning to make the model have stronger generalization and continuous learning ability.

## 2  METHOD

The overall architecture of the non-intrusive load monitoring method proposed in this paper is shown in Figure 1, which is divided into two stages: offline training and online identification. The following will describe each part of the method in detail.

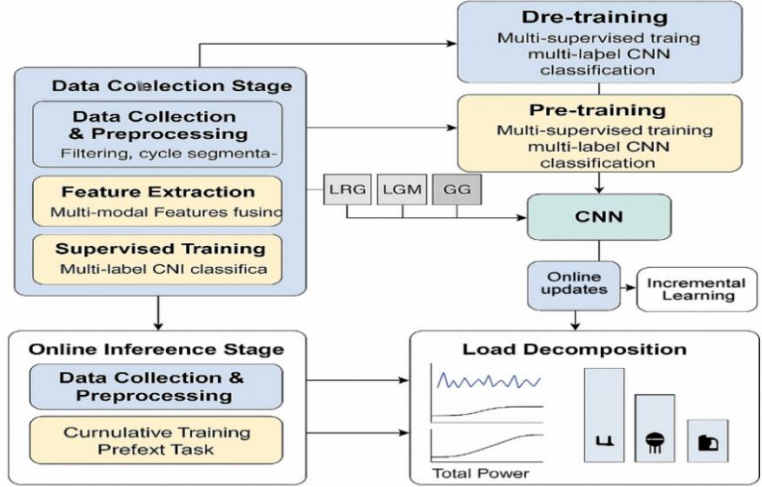

Figure 1.  Overall architecture schematic diagram of the non-intrusive load monitoring method.

## 2.1 Data collection and preprocessing

**Signal acquisition:** This study uses multiple sensors to synchronously collect current, voltage and power factor signals with sampling frequency of 50 kHz to ensure high-resolution capture of detailed changes within each cycle.

**Denoising filter:** The collected original signal is denoised to improve the signal quality. This paper adopts two-stage filtering: (1) **The low-pass filter** removes the high-frequency noise in the 50 kHz sampling signal to retain the main power frequency. The impulse response function of the low-pass filter is recorded as h(t). When the input signal is x(t), the filter output can be expressed as the convolution y(t) = (x * h)(t). By reasonably selecting the cutoff frequency (such as 1 kHz), high-frequency interference can be filtered out. (2) **The sliding mean filter** further smoothes the signal sequence. A sliding window of length N is selected, and the data in the window is averaged as the filtering result. For the sequence {x1,x2, …,xT}, the mean filter output is $x't = \frac{1}{N}\sum k = 0^{N-1} x_{t-k}$.

**Cycle division and normalization:** To extract stable load characteristics, it's necessary to divide the continuous data according to the power frequency cycle. The paper determines the cycle boundary by detecting the zero-crossing point of the voltage signal, thereby intercepting the complete single-cycle waveform data. For a 50Hz power grid, the theoretical duration of a cycle is T=1/50=0.02s. In the implementation, the starting point of each cycle can be accurately located by traversing the sampling sequence to find the point where the voltage changes from negative to positive. The corresponding current, voltage and power factor signals in a cycle are recorded as $I_{\text{cyc}}(t), V_{\text{cyc}}(t), PF_{\text{cyc}}(t)$, where $t \in [0, T]$. Periodic signals are normalized to eliminate the differences in dimensions:

$$I_{norm}(t) = \frac{I_{\text{cyc}}(t) - \mu_I}{\sigma_I}, V_{norm}(t) = \frac{V_{\text{cyc}}(t) - \mu_V}{\sigma_V}, PF_{norm}(t) = \frac{PF_{\text{cyc}}(t) - \mu_{PF}}{\sigma_{PF}} \quad (1)$$

$\mu_I, \sigma_I$ are the mean & standard deviation of a periodic current signal (the same applies to voltage and power factor).

## 2.2 Multimodal feature extraction and image load signature construction

The preprocessed current, voltage and power factor signals contain complementary load characteristic information. To make full use of the data of these three modes, this paper designs a multimodal deep feature extraction network to convert the signal into a high-dimensional feature vector and further generate a "load signature" in the form of a two-dimensional image for identification.

### 2.2.1 Multimodal time series feature extraction

We adopts a combination of temporal convolutional network and convolutional neural network architecture. Aas shown in Figure 1, for the two one-dimensional sequences of current and voltage, a TCN network module is input to capture their local time series patterns; for the scalar sequence of power factor, a one-dimensional convolution or fully connected network is used to extract its features. TCN can efficiently model long sequence dependencies [8]. Assume that the length of the $I_{norm}$ sequence is N (corresponding to the number of points in one sampling cycle), and after several layers of TCN extraction, the current feature map $F_I \in \mathbb{R}^{d_I \times N}$ is obtained, where dI is the number of current feature channels. Similarly, the voltage signal passes through another parameter-sharing TCN to obtain the feature map $F_V \in \mathbb{R}^{d_V \times N}$. For the power factor sequence, since its change is relatively gentle, this paper uses a small CNN to extract features and obtains $F_{PF} \in \mathbb{R}^{d_{PF} \times N}$. The above process is equivalent to defining three feature extraction functions $f_I(\cdot), f_V(\cdot), f_{PF}(\cdot)$, so that:

$$F_I = f_I(I_{norm}), F_V = f_V(V_{norm}), F_{PF} = f_{PF}(PF_{norm}) \quad (2)$$

The learned representations correspond to manual features but have better adaptability and discrimination.

### 2.2.2 Multimodal feature fusion

Fusion of features from current, voltage, and power factor can comprehensively utilize information from each modality to improve recognition accuracy. In this paper, $F_I, F_V, F_{PF}$ are connected in series in the feature channel dimension to form a joint feature matrix $F_{cat} = [F_I; F_V; F_{PF}] \in \mathbb{R}^{(d_I + d_V + d_{PF}) \times N}$ (the three matrices are concatenated row by row). Subsequently, after a fully connected layer is mapped to a unified dimensional space, the fused feature representation $F_{fus} \in \mathbb{R}^{d_{fus} \times N}$ is obtained, where $d_{fus}$ is the fused feature dimension. The fusion operation can be expressed as:

$$F_{fus} = \phi(W_f \cdot F_{cat} + b_f) \quad (3)$$

$W_f$ and $b_f$ are the weights and biases of the fully connected layer, and $\phi(\cdot)$ represents a nonlinear activation function. The fusion feature $F_{fus}$ integrates the time series patterns of current, voltage, and power factor.

### 2.2.3 Learnable Recurrent Graph (LRG) Features

There may be specific relationship patterns between channels in the fusion features of different appliances. In order to capture the correlation between feature dimensions, this paper introduces a **learnable recurrent graph** (LRG) representation method. In terms of specific implementation, we design a small fully connected network to map the columns of the fusion feature matrix $F_{fus}$ in pairs: for the feature vectors of the i-th and j-th columns of $F_{fus}$

(respectively denoted as $\mathbf{f}_i, \mathbf{f}_j \in \mathbb{R}^{dfus}$), we concatenate them and input them into the fully connected network and activate them through ReLU to obtain the position value $R_{ij}$ of the corresponding relationship graph matrix:

$$R_{ij} = max(0, W_r[\mathbf{f}_i; \mathbf{f}_j] + b_r) \quad (4)$$

$W_r, b_r$ are learnable parameters. By traversing all feature column pairs (i,j), we can obtain a relationship matrix $R \in \mathbb{R}^{N \times N}$, each element of which reflects the strength of the correlation between features at different times (or different phases). Figure 2 gives an example. The LRG matrices of different electrical appliances show different texture structures, which helps the subsequent CNN classifier to distinguish them.

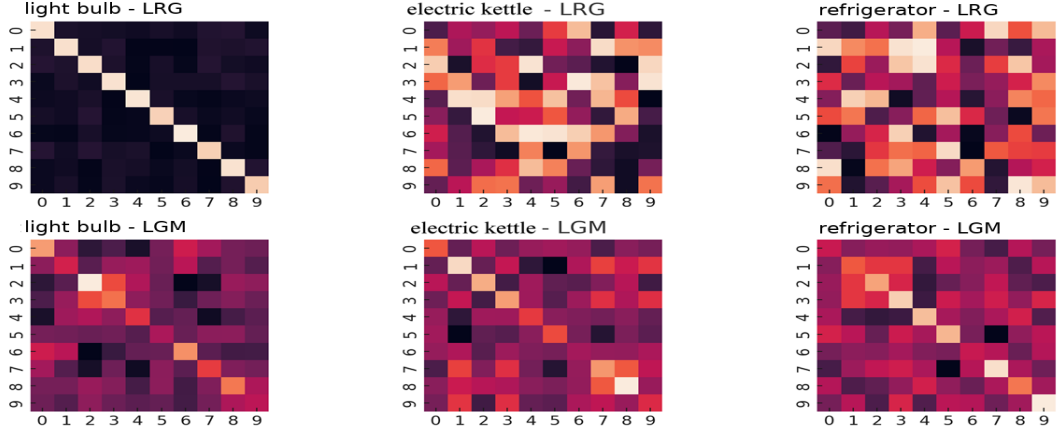

Figure 2.    Sample of Image Load Signature (LRG/LGM)

### 2.2.4 Learnable Gram Matrix (LGM) features：

The Gram matrix is a symmetric matrix obtained by calculating the inner product of a set of vectors. We introduce the idea of the Gram matrix into the representation of load features and constructs a **learnable Gramian Matrix** (LGM) feature map. Specifically, for the fusion feature matrix $F_{fus} \in \mathbb{R}^{d_{fus} \times N}$, calculate the inner product between its rows: Let $\mathbf{u}p, \mathbf{u}q \in \mathbb{R}^N$ represent the feature vectors on the pth and qth rows of $Ffus$ (i.e., the pth and qth feature channels), then the elements of the Gram matrix $G \in \mathbb{R}^{dfus \times dfus}$ are:

$$G_{pq} = \sum_{t=1}^{N} u_{p,t} u_{q,t} = \boldsymbol{u}_p \cdot \boldsymbol{u}_q \quad (5)$$

The G obtained in this way describes the similarity/correlation between different feature channels. If the value patterns of a two-dimensional feature at all time points are highly correlated, the corresponding element in its Gram matrix will be larger; otherwise, it will be smaller. The calculated Gram matrix is further transformed nonlinearly through activation functions such as ReLU to form the LGM feature map matrix $G' = \phi(G)$.

### 2.2.5 Generative Feature Map (GG)

Finally, in order to obtain a two-dimensional feature map of uniform size to input into CNN, we expand and reproject the fused features to generate a Generative Graph (GG) feature map. The specific operation is: expand $F_{fus}$ by column into a one-dimensional vector $\mathbf{z} \in \mathbb{R}^{d_{fus} \cdot N}$; then remap it back to a two-dimensional shape through a fully connected layer. For example, map $\mathbf{z}$ to a matrix of size $H \times W$ and reshape it into a two-dimensional image, where $H \cdot W$ is equal to $d_{fus} \cdot N$ (appropriate H, W can be selected as needed). The mapping can be expressed as:

$$\mathbf{z}' = W_g \mathbf{z} + b_g \quad (6)$$

and reshape $\mathbf{z}'$ into a matrix $GG \in \mathbb{R}^{H \times W}$. By learning $W_g, b_g$, the GG feature map can be trained to extract two-dimensional patterns that are beneficial to classification. The motivation for this step is that directly using the original matrix $F_{fus}$ (size $d_{fus} \times N$) as the image input may be too large to extract features with kernel.

## 2.3 Training strategies for deep learning models

After completing the above steps, each load data sample is represented as a two-dimensional feature map (or multi-channel image) X. Next, a model M needs to be trained to predict the corresponding load composition from the feature map, which is essentially a multi-label classification problem.

### 2.3.1 Self-supervised pre-training

To make use of unlabeled historical data, this paper first performs self-supervised pre-training on the feature extraction network. This study constructs a pre-training task based on **signal reconstruction**: let the model try to predict the remaining part from part of the sequence without knowing the real label, so as to learn the internal dynamic characteristics of the sequence. Specifically, given the multimodal sequence data, we intercept the first half of the cycle signal, map it to the hidden representation through the feature extraction network, and then pass it through a decoder network to predict the signal of the second half. Taking the current sequence as an example, let the first half cycle signal be $I(t), 0 \leq t < T/2$, the goal is to predict $\hat{I}(t), T/2 \leq t \leq T$, and minimize the mean square error between the predicted value and the true second half cycle I(t) during training:

$$\mathcal{L}_{ssl} = \frac{2}{T} \int_{T/2}^{T} (\hat{I}(t) - I(t))^2 dt \quad (7)$$

Similar reconstruction losses apply to voltage and power factor signals. By minimizing $\mathcal{L}_{ssl}$, the parameters of the feature extraction network are trained so that it can extract features useful for reconstructing the sequence. After several rounds of training on unlabeled data, we obtain a pre-trained model whose parameters are recorded as $\Theta_0$.

### 2.3.2 Supervised model training

Based on the pre-training parameters, we use labeled data to perform supervised fine-tuning training on the model. The model structure consists of two parts: the front-end feature extraction and the back-end CNN classifier. The front-end initialization uses the parameter $\Theta_0$ obtained by self-supervised learning, and the back-end CNN is randomly initialized. The input of the CNN classifier is the image load signature constructed above, and the output is the recognition result of each category of load. We formalize the problem as multi-label classification: Assume that there are K types of electrical appliances involved in the system. For each input sample, the model outputs a vector of length K $\hat{\mathbf{y}} = [\hat{y}_1, \hat{y}_2, \ldots, \hat{y}_K]$, where $\hat{y}_i$ represents the load state of the i-th electrical appliance predicted by the model. When $\hat{y}_i$ is close to 1, it indicates that the appliance is turned on, and when it is close to 0, it indicates that it is turned off. If power estimation is required, $\hat{y}_i$ can also be set to indicate the estimated ratio of the appliance power to the total power. The corresponding true label vector during training is $\mathbf{y} = [y_1, y_2, \ldots, y_K]$, $y_i = 1$ indicates that the i-th appliance is running in the sample (otherwise 0). **Multi-label classification cross entropy los**s is used to optimize the parameters of the CNN classifier and feature extraction network:

$$\mathcal{L}_{cls} = -\sum_{i=1}^{K} [y_i \log(\hat{y}_i) + (1 - y_i)\log(1 - \hat{y}_i)] \quad (8)$$

By minimizing $\mathcal{L}_{cls}$, the model output is made as close to the true label distribution as possible.

### 2.3.3 Continual learning and online updating

If the model only relies on offline training, its performance will decline when facing new data. Therefore, this paper introduces continual online learning strategy so that the model can learn new data during operation. Specifically, we use the Elastic Weight Consolidation incremental regularization algorithm to update model. Assuming that the model is trained on old data to obtain parameters $\Theta^*$, when a new data stream is received, we hope to impose constraints on the parameters while training the new task loss $\mathcal{L}_{new}$ so that they do not deviate from parameter values that are important for old tasks. EWC achieves this by adding a regularization term to total loss:

$$\mathcal{L}_{total} = \mathcal{L}_{new}(\Theta) + \frac{\lambda}{2}\sum_i F_i(\Theta_i - \Theta_i^*)^2 \quad (9)$$

Where $\Theta_i$ represents the i-th component in the parameter vector, $F_i$ is the importance measure of the parameter, and $\lambda$ is the trade-off coefficient. The importance measure $F_i$ is usually approximated by the Fisher information matrix calculated on the old task data:

$$F_i = \mathbb{E}_{x \sim D_{old}}\left[\left(\frac{\partial \log p(y|x; \Theta^*)}{\partial \Theta_i}\right)^2\right] \quad (10)$$

That is, the expected value of sensitivity of the parameter $\Theta_i$ to the model output probability under the parameter $\Theta^*$ trained on old task. Intuitively, the larger the $F_i$, the more important the parameter is to old task. When learning a new task, it should be kept as stable as possible without making too many changes. Through the constraints of the above regularization terms, the model will give priority to using the less important parameter capacity to learn new knowledge when training new data, thereby achieving the effect of "learning while not forgetting" [9].

### 2.3.4 Load decomposition and energy consumption calculation

After the model identifies the switch status of each appliance, it can further estimate the power and energy consumption of each appliance. We use a **variational autoencoder** (VAE) to decompose the total power signal into several independent components. When training VAE, the power curve of each appliance when running alone in the historical data is used as a priori to let the model learn to generate the power pattern of a single device; then in the inference stage, the total load power is used as input, and the most likely sum of the power of the constituent appliances is decoded through the latent variables of VAE. Formulaically, it can be expressed that the total load power sequence $P_{\text{total}}(t)$ is decomposed into the sum of the power $P_i(t)$ of K devices:

$$P_{\text{total}}(t) \approx \sum_{i=1}^{K} P_i(t) \quad (11)$$

Where $P_i(t)$ represents the power reconstruction component of the i-th type of appliance. If a certain type of appliance is identified as not running, the corresponding $P_i(t)$ is approximately 0. By integrating $P_i(t)$ over time, we can get the electric energy $E_i = \int P_i(t)dt$ consumed by the device in a certain period.

## 3 EXPERIMENTS AND RESULTS

In order to verify the effectiveness of the method proposed in this paper, we designed and conducted a large number of comparative experiments and analyses.

### 3.1 Dataset and Experimental Settings

*3.1.1 Dataset*

We train and test the model on the following datasets, which are divided in a ratio of 8:2:

(1) **PLAID dataset[4]:** a public high-sampling rate load monitoring dataset containing voltage and current waveform fragments of 11 household appliances with a sampling frequency of 30 kHz. We select the voltage and current signals and calculate the power factor sequence to construct a multi-modal input.

(2) **WHITED dataset[6]:** contains high-frequency waveforms of the on/off transients of household appliances, which is used to test the model's ability to detect the moment of device startup.

(3) **Self-collected load data:** We built a smart power consumption experimental platform and collected current, voltage, and power factor data of 8 common household appliances under different working conditions in a laboratory environment. The sampling frequency was 50 kHz to simulate the bus signal in the real scenario.

*3.1.2 Comparative Methods*

In order to comprehensively measure the performance of our method, we selected the following representative baselines for comparison:

1) **Traditional feature + classifier method:** Extract classic features such as steady-state power and current peak, and train support vector machine (SVM) and random forest classifier for load classification respectively [6]. This represents the traditional NILM method without deep learning.

2) **V-I trajectory image + CNN method [3]:** Use the voltage-current trajectory of each cycle to construct a grayscale image, and then train a convolutional network (ResNet-18) for classification. This method represents the current popular load "fingerprint" image recognition idea.

3) **Time series deep learning method:** The total current sequence is directly input into the sequence model, and the power estimation of each device is output (i.e. sequence-to-sequence or sequence-to-point regression) [2].

*3.1.3 Evaluation Metrics*

For the load classification results, we use indicators such as Accuracy, Precision, Recall etc. The accuracy in the multi-label case is defined as the proportion of running devices that the model correctly identifies in samples; the precision and recall are calculated for each type of device and macro-averaged. For the accuracy of decomposition, we use the mean absolute error to measure the difference between the power estimate and the true value..

**3.2 Performance comparison results**

*3.2.1 Overall recognition accuracy*

Table 1 shows the main indicators of all methods on test set. Traditional ML methods have only an accuracy of 60% due to simple features, and are difficult to deal with mixed load situations [1]. The V-I trajectory image combined with CNN method achieved an accuracy of 85%, indicating that image signatures have advantages in load identification. The sequence-to-point DL method achieved an accuracy of 88% and an average F1 score of 0.90. In contrast, the proposed method combines multimodal features, self-supervised pre-training, and continual learning, achieved the highest overall accuracy of 92.5%, an average F1 score of 0.93, and achieved an F1 value of more than 90% on each type of device. Especially in samples with multiple devices at the same time, the multi-label recognition accuracy of the proposed method is significantly better than other methods, reflecting the effectiveness of fusion features and image signatures in distinguishing complex load combinations.

Table 1. Comparison of overall recognition performance of different methods

| Method | Accuracy | Precision | Recall | F1-score |
| --- | --- | --- | --- | --- |
| Traditional features +SVM | 0.603 | 0.612 | 0.599 | 0.605 |
| CNN+ V-I trajectory image | 0.851 | 0.864 | 0.843 | 0.85 |
| Sequence CNN | 0.879 | 0.891 | 0.886 | 0.888 |
| Proposed | 0.925 | 0.933 | 0.93 | 0.931 |

*3.2.2 Segment performance*

Figure 3 further compares the F1 scores of different methods on various appliance categories. The proposed method has achieved leading recognition accuracy on various types of devices, verifying the effectiveness of the method.

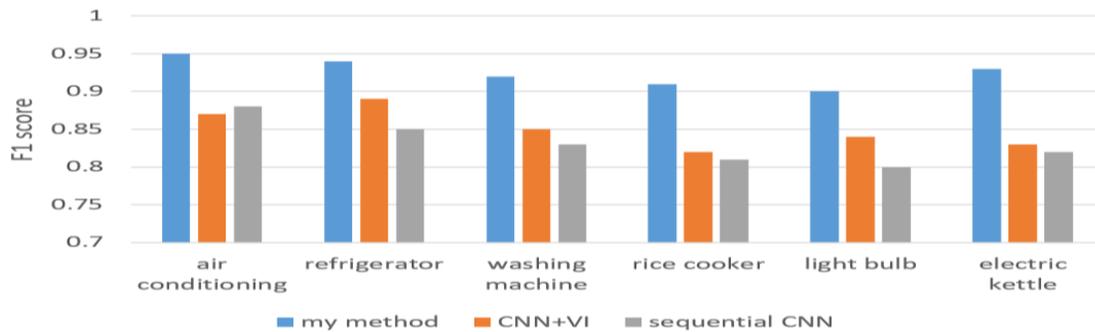

Figure 3. Comparison chart of F1 scores of models in various categories of loads

## 4 CONCLUSION

Focusing on the key issue of non-intrusive load identification, we proposes an innovative method based on image load signature and continual learning. Based on an in-depth analysis of the limitations of existing technologies, we introduce multimodal high-frequency signal fusion and visual signature features, and use deep convolutional networks to accurately classify complex load combinations. Meanwhile, we alleviate insufficient data through self-supervised pre-training, and use the elastic weight retention algorithm to continuously learn new loads.

**ACKNOWLEDGMENTS**

This work Supported by Smart Grid-National Science and Technology Major Project 2024ZD0802100.

**REFERENCES**

[1] Du, Z., Yin, B., Zhu, Y., Huang, X., & Xu, J. **2023**. A NILM load identification method based on structured VI mapping. Scientific Reports, 13(1), 21276.

[2] Chen, S., Zhao, B., Zhong, M., Luan, W., & Yu, Y. **2023.** Nonintrusive load monitoring based on self-supervised learning. IEEE Transactions on Instrumentation and Measurement, 72, 1-13.

[3] Shi, C., Gan, L., Yang, T., Zhang, P., Yu, K., & Mei, F. **2024, October**. Non-Intrusive Load Monitoring Method Based on Color-Coded VI Trajectories and Multi-Feature Fusion. In 2024 IEEE PES 16th Asia-Pacific Power and Energy Engineering Conference (APPEEC) (pp. 1-5). IEEE.

[4] Lin, L., Zhang, J., Gao, X., Shi, J., Chen, C., & Huang, N. **2023**. Power fingerprint identification based on the improved VI trajectory with color encoding and transferred CBAM-ResNet. PloS one, 18(2), e0281482.

[5] Sayed, A. N., Himeur, Y., Varlamis, I., & Bensaali, F. 2025. Continual learning for energy management systems: A review of methods and applications, and a case study. Applied Energy, 384, 125458.

[6] Lu, L., Kang, J. S., Meng, F., & Yu, M. **2024**. Non-intrusive load identification based on retrainable siamese network. Sensors, 24(8), 2562.

[7] Paknezhad, M., Rengarajan, H., Yuan, C., Suresh, S., Gupta, M., Ramasamy, S., & Lee, H. K. 2023. Improving transparency and representational generalizability through parallel continual learning. Neural Networks, 161, 449–465.

[8] Chen, T., Gao, J., Yuan, Y., Guo, S., & Yang, P. **2024**. Non-intrusive load monitoring based on MoCo_v2, time series self-supervised learning. Energy and Buildings, 317, 114374.

[9] Kirkpatrick, J., Pascanu, R., Rabinowitz, N., Veness, J., Desjardins, G., Rusu, A. A., ... & Hadsell, R. **2017**. Overcoming catastrophic forgetting in neural networks. Proceedings of the national academy of sciences, 114(13), 3521-3526.